\documentclass{article}

\usepackage{microtype}
\usepackage{graphicx}
\usepackage{subfigure}
\usepackage{booktabs} 
\usepackage{hyperref}

\usepackage[accepted]{icml2021}

\usepackage{amsmath}
\usepackage{amssymb}

\usepackage[normalem]{ulem}
\useunder{\uline}{\ul}{}

\begin{document}
\twocolumn[
\icmltitle{Enhancing Privacy against Inversion Attacks in Federated Learning by using Mixing Gradients Strategies}





\icmlsetsymbol{equal}{*}
\begin{icmlauthorlist}
\icmlauthor{Shaltiel Eloul}{goo}
\icmlauthor{Fran Silavong}{goo}
\icmlauthor{Sanket Kamthe}{goo}
\icmlauthor{Antonios Georgiadis}{goo}
\icmlauthor{Sean J. Moran}{goo}

$^{1}$CTO, JPMorgan Chase, London, UK. 
\vskip 0.3in
\end{icmlauthorlist}








\begin{abstract}
Federated learning reduces the risk of information leakage, but remains vulnerable to attacks. We investigate how several neural network design decisions can defend against gradients inversion attacks. We show that \emph{overlapping} gradients provides numerical resistance to gradient inversion on the highly vulnerable dense layer. Specifically, we propose to leverage batching to maximise mixing of gradients by choosing an appropriate loss function and drawing identical labels. We show that otherwise it is possible to directly recover all vectors in a mini-batch without any numerical optimisation due to the de-mixing nature of the cross entropy loss. To accurately assess data recovery, we introduce an \emph{absolute variation distance} (AVD) metric for information leakage in images, derived from total variation. In contrast to standard metrics, e.g. Mean Squared Error or Structural Similarity Index, AVD offers a continuous metric for extracting information in noisy images. Finally, our empirical results on information recovery from various inversion attacks and training performance supports our defense strategies. These strategies are also shown to be useful for deep convolutional neural networks such as LeNET for image recognition. We hope that this study will help guide the development of further strategies that achieve a trustful federation policy.
\end{abstract}
]

\section{Introduction}
Federated learning (FL) enables distributed client nodes to contribute to the training of a centralised global model without exposing their private data~\cite{pmlr-v54-mcmahan17a,Kairouz19,Yang19}. The promises of federated learning are significant and have wide applicability in industry. For example through federated learning it is possible for hospitals to collaborate on training a centralised model around the globe, without sharing or moving the actual private patient information across institutions~\cite{Rieke20}. As it potentially protects sensitive data, it can better align with data protection regulations such as GDPR~\cite{Commission18}. For example, FL has been already applied to prediction of treatment side effects in medicine~\cite{Jochems2016DistributedLD} or learning a predictive keyboard for smartphones~\cite{Bonawitz19,Konecny16}.
The reduction of data movement is an additional important advantage, as it is costly and time consuming for large industrial applications. 
Given the potential impact of FL, many authors have since examined the security and privacy of FL~\cite{zhao2020idlg,GeipingBD020,Yin21,Zhu19,Huang20,Phong18,Carlini20,ShokriSSS17,MelisSCS19}.
A standard FL configuration is typically achieved with a central aggregator node which exchanges gradients for centralised aggregation. At each training step ($t$), a client node receives neural network model weights, $F(W_t)$, from the aggregator server and calculates loss ($l$) with a local data $x_t,y_t$ for a mini-batch, $B$, which generates gradients with respect to the model weights:
\begin{equation}
\Delta W_t=-\frac{\gamma}{B} \sum_{b<B}{\frac{\partial l(F(x_{t,b},y_{t,b}))}{\partial W_t}}
\end{equation}
 
The gradients are aggregated by the centralised server, usually by averaging and with rate, $\gamma$. The gradients, $\Delta W_t$, shared by the client can expose the client to a potential inversion attack instigated by a malicious eavesdropper. The inversion attacks have shown to be surprisingly successful in many pioneer studies~\cite{zhao2020idlg,GeipingBD020,Zhu19}. This compromised privacy prevents federated learning from becoming a fully trustful framework for distributed training of machine learning models. Whilst most prior studies are focused on developing successful inversion attacks on the gradients, in this work, we identify conditions under which privacy can be attained with confidence against inversion attacks. We explore the effect of the objective loss function and its relation to the mini-batch class distributions for dense layers. In addition, we also conduct a similar study for convolutional neural networks on the benchmark task of image classification. Through our analysis, we demonstrate that \emph{maximising the mixing of gradients} can be an effective defense strategy to counteract gradient inversion attacks. Furthermore this defense strategy does not require or minimise the need for masking of the gradients by pruning or noise addition which often leads to a degradation on model accuracy~\cite{Zhu19,wei2020framework,Tramer21}. In more detail, our contributions in this paper are three-fold: 

\begin{itemize}
\item \textbf{Inversion of batch, label distribution and loss function:} we revisit the linear dense layer and show that it is possible to directly invert a full batch without an optimisation-based gradient attack. In this context, we provide insights into the role of the loss function and batch label distribution on the recovery rate of gradient inversion attacks. 
\item \textbf{Strategies for better privacy:} resulting from the above study, we suggest and provide a quantitative analysis of training strategies that are stable against attack. In contrast to existing defence mechanisms (e.g. differential privacy), our training strategies require no noise, therefore preserving accuracy in classification tasks with no performance degradation.
\item \textbf{Absolute Variation Distance (AVD):}
it is usually difficult to tell automatically whether information leakage has obtained from a gradient reversal attack, and metrics such as mean squared error between the ground-truth and recovered data are inadequate due to the high degree of noise. We introduce a more effective metric for detecting information recovery that we term as absolute variation distance, a variant of total variation metric~\cite{Rudin92}. This metric can have an important applications for creating a policy as an automatic tool for continuous monitoring of information leakage in a federation, and in future studies of defence strategies for FL. 
\end{itemize}

\section{Related Work}\label{sec:related_work}
There is a wide literature of attacks on machine learning models, and a significant portion of recent work focuses on comprising deep learning models to steal private data~\cite{Zhu19} and even hyperparameters~\cite{WangG18}. Our study is mostly related to attacks on learning systems that run on multiple distributed nodes and exchange gradients to communicate weight updates. 
Despite the compelling promises for privacy in FL, there is a body of work~\cite{zhao2020idlg, GeipingBD020, Yin21, Zhu19, MelisSCS19, ShokriSSS17} that present eavesdropping attacks on distributed machine learning systems to compromise data privacy, necessitating a better understanding on defense mechanisms to generate a trustful federation policy.


\noindent \textbf{Gradient Inversion Attacks:}
Early works studied techniques for extracting metadata about the private data, for example, membership attacks have been proposed in which a classifier is trained to identify whether a specific data-point has been used to train a model~\cite{ShokriSSS17}. Property attacks are another attack variant in this direction where properties of a batch are exposed~\cite{MelisSCS19}, such as the presence of a person in a photograph or their age. In both cases, actual data-points are not extracted from the gradient information. Later studies, led by Zhu~\emph{et al.}~\cite{Zhu19} and expanded upon by~\cite{zhao2020idlg,GeipingBD020,Yin21,ZhuB21} showed how it is possible to extract the actual data by inverting the gradients communicated by clients in a federation. For example, recently~\citet{Yin21} showed that it is possible to extract data at a pixel-level granularity with remarkable clarity. Many approaches tackle the inversion as an optimisation problem, and we discuss the popular algorithms in the context of our experiments in section~\ref{sec:optim_attacks}. 


Much less attention is dedicated for defence of those inversion attacks. However typical mechanism has been proposed with a range of effectiveness, including gradient pruning,  adding noise to the gradients~\cite{Zhu19,wei2020framework,Tramer21} and encryption of training images applied online by the client nodes~\cite{Huang20}\footnote{Recently a input defence scheme has been proposed for InstaHide~\cite{Carlini20}  }.

In this paper we propose a defence strategies that leverage the properties of a mini-batch to maximise gradient mixing. We analyse \emph{gradient mixing} as a lightweight defense strategy for counteracting gradient inversion attacks. 

\section{Gradient mixing as a batch defense strategy}
We focus in the analysis of gradient mixing on a mini-batch of linear dense layers. Later in the paper, we provide supporting results for classification tasks of linear-layer and LeNET.
\subsection{Batch size of a final dense layer}
The high vulnerability for recovering vector information from a fully connected layer is well known~\cite{zhao2020idlg,qian2021minimal,GeipingBD020}, but here we examine the vulnerability of averaged information from a batch and under various gradient mixing conditions. We simplify our analysis to a dense linear layer containing only $x$ as input and $y$ as output where, $o_j=\sum^n_i{w_{ij}x_i+b_j}$. Note that $x$ can be inverted from known $y$, and any additional hidden linear layers can be inverted by back-propagation (however, this is not the case for convolutional layers where inputs are convolved by shared kernel weights).  

\subsection{Direct inversion of a full batch}
A typical classification architecture uses softmax, $p_k=\frac{e^{o_k}}{\sum_j{e^{o_j}}}$, followed by cross-entropy to obtain the loss:
\begin{equation}
  l(p,y)=-\sum_k^C{y_k \log p_k}
\end{equation}where, $C$ is the number of classes/categories. The derivative of $p_k$ with respect to each $o_j$:
\begin{equation}
  \frac{\partial{p_k}}{\partial{o_j}}=\begin{cases}
    p_k(1-p_j), & k=j \\
    -p_k p_j, & k \neq j.
  \end{cases}
\end{equation}
The loss set of equations is then obtained:
\begin{equation}
\frac{\partial{l}}{\partial w_{i,j=k}}=\frac{\partial{l}}{\partial{p_k}}\frac{\partial{p_k}}{\partial{o_j}}x_i=(p_j-y_j)x_i
\label{eqW1}
\end{equation}
and:
\begin{equation}
\frac{\partial{l}}{\partial b_{j=k}}=p_j-y_j
\label{eqB1}
\end{equation}
Specifically, the gradients are shared accurately for a single input batch, $B=1$, as also observed by \citet{qian2021minimal}.
The number of gradient equations are $nC+C$ with an extra $C$ equations for weights (for each $j$) whilst the unknowns are $n+C$. For example $x_i$ can be found from any $j$, using $\frac{\partial{l}}{\partial w_{ij=k}}/\frac{\partial{l}}{\partial b_{j=k}}= x_i$. 
Although this shows the vulnerability of a linear layers in an FL model, when the mini-batch size is larger than one, the client would only share the averaged information:
\begin{equation}
\frac{\partial{l^B}}{\partial w_{i,j=k}^B}=\frac{1}{B}\sum_{m \in [1,B]}{(p^m_j-y^m_j)x_i^m}
\label{eq2W1}
\end{equation}

\begin{equation}
\frac{\partial{l^B}}{\partial b^B_{j=k}}=\frac{1}{B}\sum_{m \in [1,B]}{p^m_j-y^m_j}
\label{eq2B1}
\end{equation}

As no additional equations are shared, the number of unknown, $B(n+C)$ can exceed the gradients equations number $nC+C$, and there will be no unique solution to solve the set of equations. Even in the case that a unique solution exists, numerical optimisation can be challenging. However, in the scenario in which softmax is followed by cross entropy, we show that an accurate direct solution can be found in many cases even for $B \gg 1$, due to the de-mixing property across the batch.
In an untrained, randomised weights model, the first order expected value of $\langle p^m_j-y^m_j \rangle$, is positive but close to zero for a non-target instance ($j\neq c$) and close to $-1$ for the instance target ($j=c$). This is due to the fact that expected value, $\langle p_j(o_j) \rangle$ can be grossly estimated as $p_j(E(o_j))$ for first order Taylor expansion (see also \cite{daunizeau2017semianalytical}). Therefore, this results in $\langle p_j \rangle$ to be inversely proportional to the number of classes $C$.
Subsequently, a batch that contains non-similar labels, $c^m \neq c^{1 ... B}$,  Eq.~\eqref{eq2W1}-\eqref{eq2B1} can be estimated as:
\begin{equation*}
\frac{\partial{w_{i,j}^B}}{\partial b^B_{j}}\approx \frac{(\langle p^m_{j=c} \rangle-1)x_i^{m(j=c)}+\langle p^m_{j \neq c} \rangle \sum x_i^{m(j \neq c)}}
{(\langle p^m_{j=c} \rangle-1)+(B-1)\langle p^m_{j \neq c} \rangle}
\end{equation*}
\begin{equation}
\approx \frac{(\langle p^m_{j=c} \rangle-1)x_i^{m(j=c)}}{(\langle p^m_{j=c} \rangle-1)}=x_i^{m(j=c)}
\label{eq:batch_estimate}
\end{equation}

This approximation shows that the de-mixing of the gradients for each vector enables a direct estimate of the input layer for any vector $m$ in the batch when we pick $j=c$. The error of this estimation can be very low for large $C$. It is clearly seen empirically with the two most popular data-sets used for studying such inversion attacks, the MNIST with small number of equations $C$ and LFW with large $C$ ($C \gg B$ ).

\begin{figure}[t!]     
\subfigure[]{\includegraphics[trim={0.8cm 0.2cm 0.9cm 0.5cm}, width=0.22\textwidth]{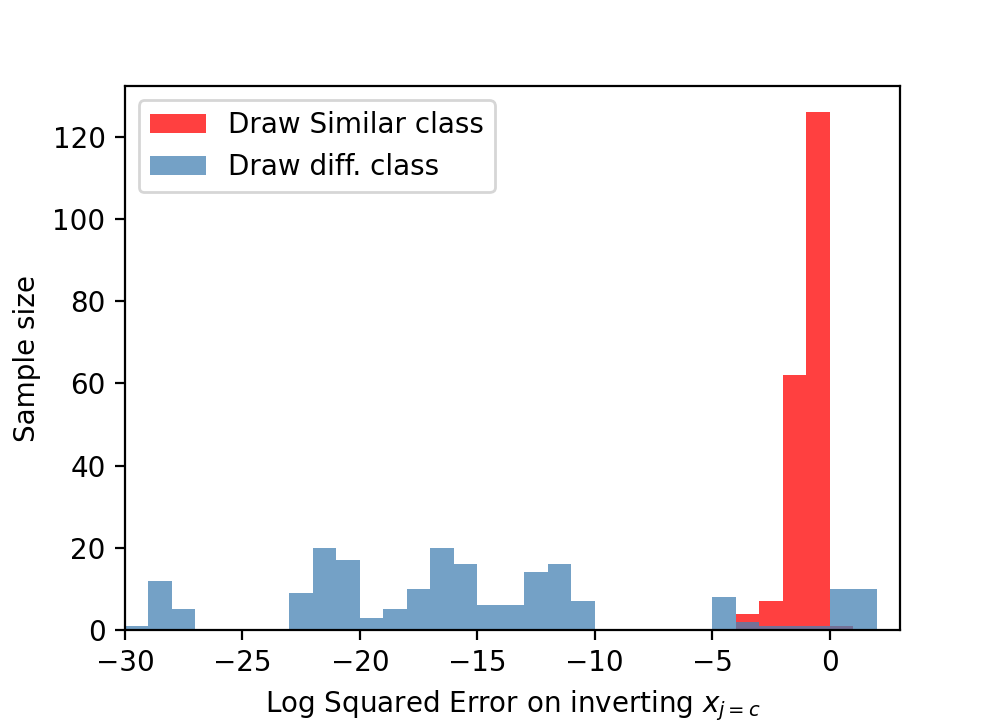}}
\subfigure[]{\includegraphics[trim={0.8cm 0.2cm 0.9cm 0.5cm},width=0.22\textwidth]{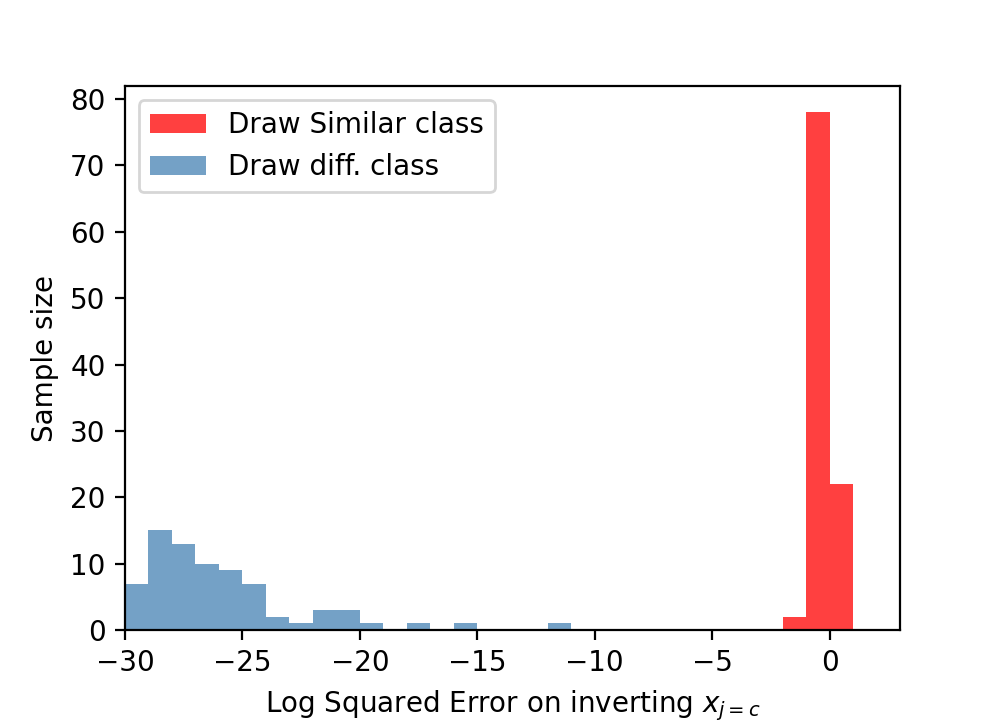}}
\centering
\subfigure[]{\includegraphics[trim={0.8cm 0.2cm 0.9cm 0.5cm},width=0.25\textwidth]{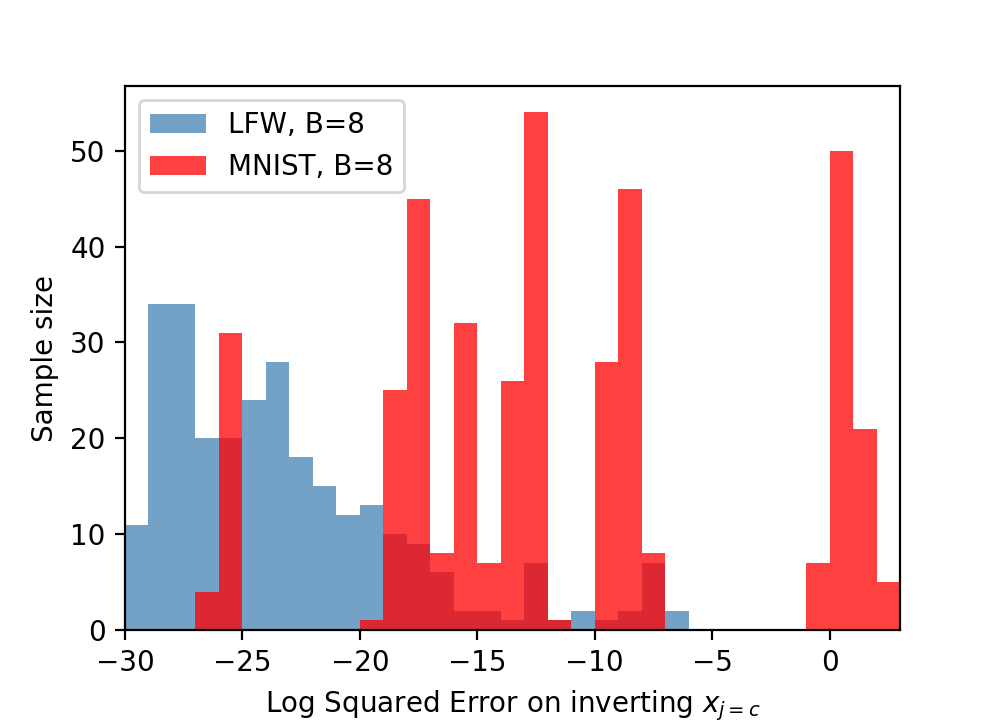}}
\caption{Log error on direct inversion on input batch for linear dense layer. We calculate the approximation of Eq.~\eqref{eq:batch_estimate} in 2 histograms for an untrained model. The case of drawing similar labels (red), and drawing unique labels in the batch (blue). (a) for MNIST (C=10), (b) for LFW dataset (C=5749), and (c) for batch size of 8 drawing unique labels, comparing LFW (blue) and MNIST (red).}
\label{histo-labels}
\end{figure}

Figure~\ref{histo-labels}(a)-(c) shows our estimates to direct invert the input of a dense layer from a vector batch, and the ability to infer all inputs of a batch in a dense layer. It supports that inverting the 2-8 vectors is likely to be possible with a very small error, as long as labels are unique. In the case of the LFW dataset, due to the large $C$, the error is inversely proportional to $C$, so the two vectors in the batch are recovered for all random samples tested with a very low error. This low error is obtained so long as the vectors have a unique class, given $C \gg B$, which is the case for the LFW classification network with much larger batch sizes (e.g. $B \gg 8)$.

The de-mixing property of cross-entropy is not only helpful for estimating input without numerical optimisation, but also allows simple numerical convergence as the dimensionality for searching a solution is effectively reduced.  This can be inferred from Eq.~\eqref{eq:batch_estimate}. If for a given $j=c$, $x_i$ only belongs to a single $m=M$, the search is constrained to a one dimension parabola (in the case of a squared distance objective function), as other $m \neq M$ will not influence the right term in Eq.~\eqref{eq:batch_estimate}. Yet, once we draw similar labels in the batch the recovery of data exhibits a large error, and the error is within the order of magnitude of the information ($\propto x_i$). This mixing of gradients can serve as a strategy to increase privacy in FL against direct inversion and our results in section~\ref{sec:optim_attacks} show that the strategy is also effective against numerical optimisation attacks. 

\begin{figure}[ht]
\subfigure[]{\includegraphics[trim={0.65cm 0.0cm 0.9cm 0cm},width=0.235\textwidth]{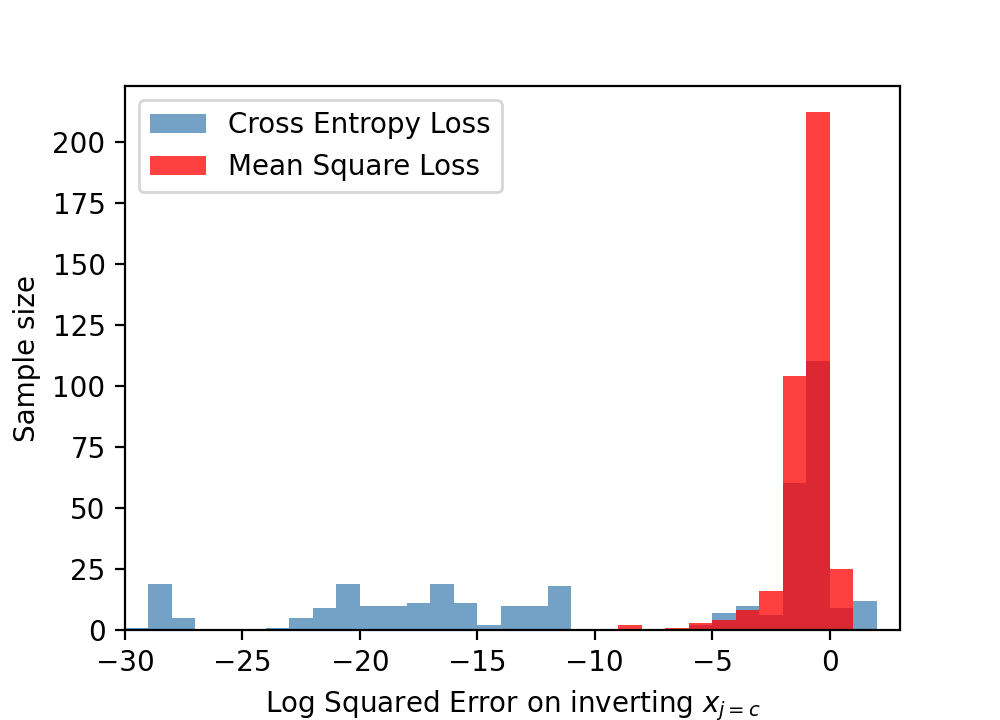}}
\subfigure[]{\includegraphics[trim={0.65cm 0.0cm 0.9cm 0cm},width=0.235\textwidth]{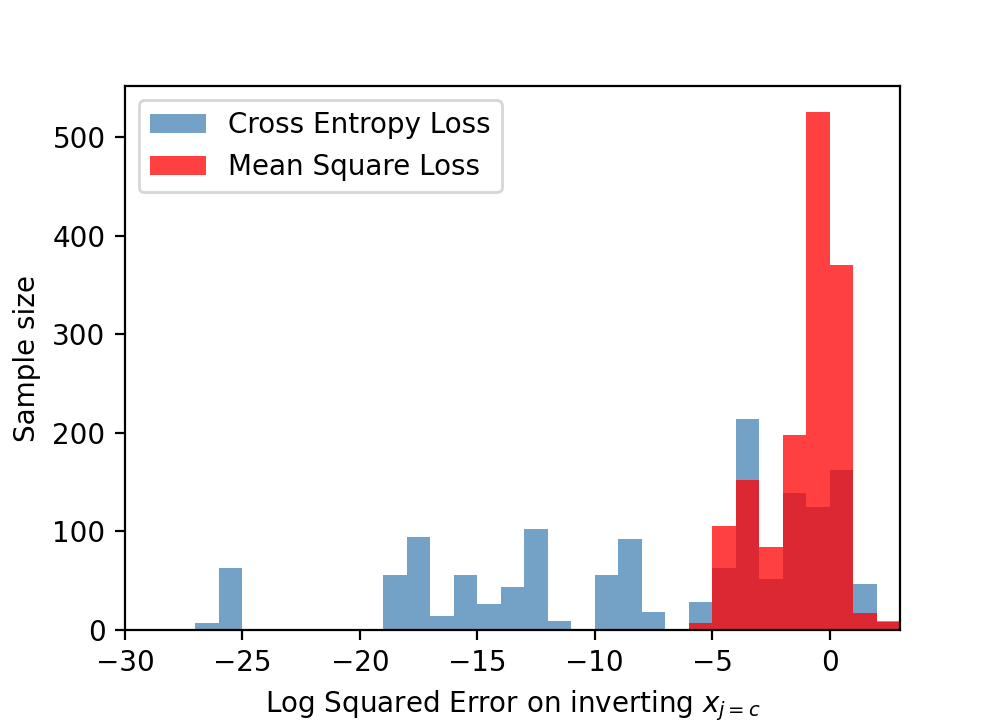}}

\caption{The error on direct inversion of single input in a linear dense layer in a batch. We calculate the approximation in Eq.~\eqref{eq:batch_estimate} and show the error in two histograms for an untrained model. The case of mean square error objective loss function (red histogram) compared to the cross entropy loss as generally used (both use softmax). Histogram (a) shows the result of minibatch size of 2, and histogram (b) for batch size of 8.}
\label{loss-histogram}
\vskip -0.2in
\end{figure}

Following this insight, we can also consider changing the objective function to mix the gradients. Instead of using cross-entropy loss (CEL), it is possible to use the mean squared loss (MSE), $l_2(o,y)=-\sum_k^C{(y_k-o_k)}^2$. This is not a typical choice for a classification task, but performance results show later that there is little to an unnoticeable degradation in classification performance when using MSE instead of CEL. On the contrary, there is a large gain in privacy by the high mixing of gradients on the dense layer. The gradients are calculated on a dense layer using mean square estimation (for simplicity with no softmax):
\begin{equation}\label{eqn:mse_est_w}
\frac{\partial{l_2}}{\partial w_{i,j=k}}=\frac{\partial{l_2}}{\partial{o_k}}\frac{\partial{o_k}}{\partial{w_i}}=-2(o_j-y_j)x_i
\end{equation}
and
\begin{equation}\label{eqn:mse_est_b}
\frac{\partial{l_2}}{\partial b_{j=k}}=-2(o_j-y_j)
\end{equation}
Here $o_j$ can take positive or negative values and generally within the order of $y$, for $j=c$ or $j \neq c$, hence the average of a batch will not de-mix the gradients.  Note that we are also using softmax followed by MSE for comparison, and the gradients will also obtain a strong gradient mixing (but a less trivial gradient expression).
We show in figure~\ref{loss-histogram} a sample of results for the error on the direct inversion of a batch. We observe that the error on estimation of any vector using $\frac{\partial w}{ \partial b}$ is not negligible, and sufficiently large even in batch size of 2 and more distinctive in batch size of 8.

We see here that potentially drawing input with similar labels and adjusting the loss function to increase mixing of gradients are strategies that counter the recovery of input from dense layers. In fact, we can compare our result to the effectiveness of adding noise as a defensive strategy given its wide application. We add a Gaussian noise term for all gradients in a linear layer, $\frac{\partial{l}}{\partial w_{i,j}} +\zeta_{i,j},
\frac{\partial{l}}{\partial b_{j}} +\zeta_j$.

\begin{figure}[ht]
\subfigure[]{\includegraphics[trim={0.65cm 0.0cm 0.9cm 0cm},width=0.235\textwidth]{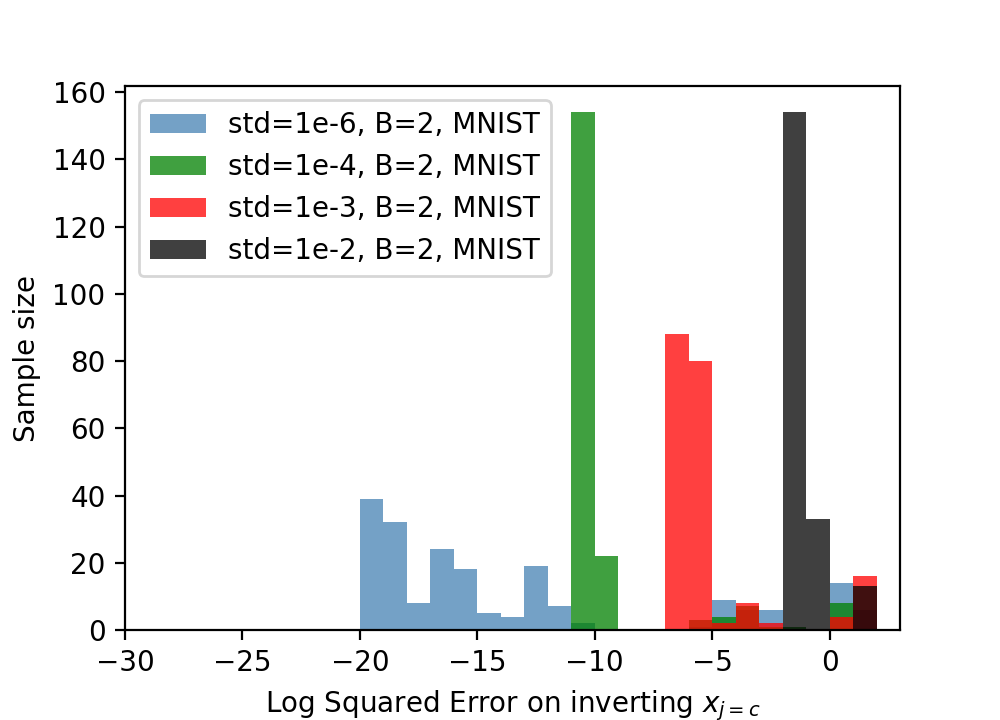}}
\subfigure[]{\includegraphics[trim={0.65cm 0.0cm 0.9cm 0cm},width=0.235\textwidth]{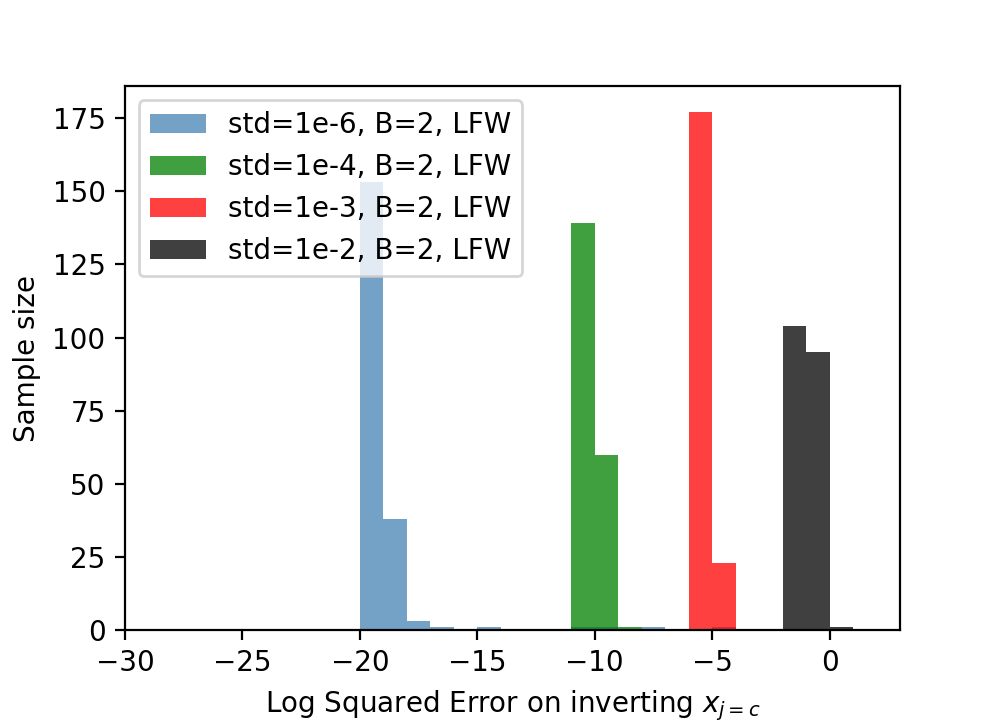}}

\caption{The error from direct inversion by adding Gaussian noise to gradients and biases at various standard deviations. Histogram (a) for MNIST dataset and histogram (b) for LFW dataset.}
\label{noise-histogram}
\vskip -0.2in
\end{figure}

Figure~\ref{noise-histogram} shows the results from addition of noise at various standard deviations. We find that small contamination of noise does not protect against inversion and the error is in the order of the noise, and given that, we are required the noise to be in the order of the weights ($std>0.01$, with weights initialised uniformly between $(-0.5,0.5)$). The addition of such a noise will affect the training drastically. In fact, this exercise shows that our gradient mixing strategies can be as effective as the addition of a large noise term, without the loss of training performance as we show in section~\ref{sec:results}.

To further support our gradient mixing strategies, in section~\ref{sec:results} we carry out an analysis of widely used inversion attacks that use numerical optimisation. We do this by firstly analysing inversion attack success in a dense layer model for simplicity, and then show its validity for a typically explored convolutional network in an inversion attack context.

\begin{table*}[htp]
\caption{Types of gradient inversion attacks employed to evaluate our proposed defense strategy.}
\resizebox{\textwidth}{!}{\begin{tabular}{llll}
\hline
{Attack Name} & {Main Objective Function} & {Description} \\ \hline
2-norm & $g^{l2}$ \eqref{eqn:gl2} & Euclidean distance and initial label determination. \\
Angle \& var  & $g^{ang}$ + TV \eqref{eqn:gang} & \citet{GeipingBD020} proposed to leverage cosine 
similarity, total variation (TV) and initial label determination. \\
Angle \& var \& Orth\_regulators & $g^{ang}$ + TV + Orth  & Cosine distance with orthogonal regulator for the input +  initial label determination. \cite{qian2021minimal}\\
\\ \hline
\label{attacks}
\end{tabular}}
\end{table*}

\section{Experimental framework}
In this section we detail the experimental framework we use to evaluate our gradient de-mixing strategies in section~\ref{sec:results}. In our experiments we explore the privacy of the input data with two representative networks, the first, a single dense layer and a standard LeNET convolutional neural network~\cite{Lecun90}. We analyse the impact of different loss functions and label distributions by varying the batch sizes and the number of filters. This empirical study explores the limit of the attacks for varying conditions. For example, how many filters and the size of batch that are needed to protect privacy. We demonstrate how the MSE loss and choosing similar labels in a mini-batch helps a practitioner leverage a greater number of filters and a lower size of batch. 

\subsection{Inversion Attack Optimisation Algorithms}\label{sec:optim_attacks}
The gradient inversion attack is carried out by choosing $x_t',y_t'$ on a proxy model, $F'(x_t',y_t')$, and finding $\Delta W_t'$ which minimises an objective function $M\Delta W_t$. A typical objective can be the norm of the gradients' difference:
\begin{equation}\label{eqn:gl2}
  g^{l2}(x_t', y_t')=\mathrm{min} ||\Delta W_t'-\Delta W_t||
\end{equation}
This solution searches for a model $F'(x_t',y_t')$ that matches the size of the gradient vector observed by the client. Although further empirical studies have found the cosine distance to provide better convergence results~\cite{GeipingBD020}:
\begin{equation}\label{eqn:gang}
  g^{ang}(x_t', y_t')=\mathrm{min}  \, 1 - \frac{\langle \Delta W_t',\Delta W_t \rangle}{||\Delta W_t'||\cdot||\Delta W_t||}
\end{equation}
Various regularisation terms were shown to improve convergence. For example, regularisation that penalises high variations in the input images and constrains the search to high-fidelity images and de-noised solutions~\cite{GeipingBD020,Yin21}. In mini-batches the orthogonality \cite{qian2021minimal} between input vectors in the batch has been shown to bias the search towards different vectors in the batch. Additionally it has been found that determining the label from the gradients is important for initialisation of the numerical optimisation~\cite{Yin21}.
We have also seen in the literature various type of attacks that provide improvements in image fidelity, or training convergence. Since no work so far is focused on enforcing the leakage of minimal information, in our study in section~\ref{sec:results} we apply various types of attacks and regularisation terms to provide a comprehensive analysis without any prior assumption on the performance of the attack. As summarised in Table~\ref{attacks}, we utilise both the Euclidean distance and cosine similarity objective functions proposed by recent prior work~\cite{Zhu19,GeipingBD020} including a selection of popular regularisation functions.

\subsubsection{Determining the Labels in a Mini-Batch}
One of the important characteristics for fast attack convergence is the initial determination of the labels which reduces the number of unknowns in the dense layer from $nB+CB$ to only $nB$ (for a classification task only). A typical classification model contains a softmax followed by a cross-entropy loss on neural network output. Labels can be obtained in this context by choosing the negative gradients to be the labels. Recall if $p_j^m$ is close to zero in an untrained model, from the distribution of the negative gradients it is possible to identify all the distribution of labels and even similar labels, as has also been suggested previously \cite{zhao2020idlg,Yin21}. However here we stress that this is possible only when $C \gg B$ but not necessarily true when $B$ is comparable to $C$. This is due to the fact that $p_j^m$ (see Eq.~\eqref{eqW1}) can obtain significant positive values for certain $j$ when $C$ is small, and can even result in average positive gradients when $j=C$. This for example occurs in the MNIST dataset where $C=10$ but will not be observed in LFW with large $C$. In the case of the MSE loss, we cannot initialise the labels distribution directly from the gradient sign, as $o_j$ is proportional to $y_j$ in Eq.~\eqref{eqn:mse_est_w} and Eq.~\eqref{eqn:mse_est_b} . Hence, we suggest that although we determine the overall label distribution, we find that it is important to still optimize the labels output using the optimizer scheme in order to maximise the recovery of an MNIST attack for a batch.
Note that in a well trained model, it is also possible to estimate labels of a batch directly from $p_j^m$, so typically FL frameworks do not protecting labels.

\begin{figure}[t!]  
\subfigure[]{
\includegraphics[trim={0.9cm 0.2cm 0.0cm 0cm},width=0.49\textwidth]{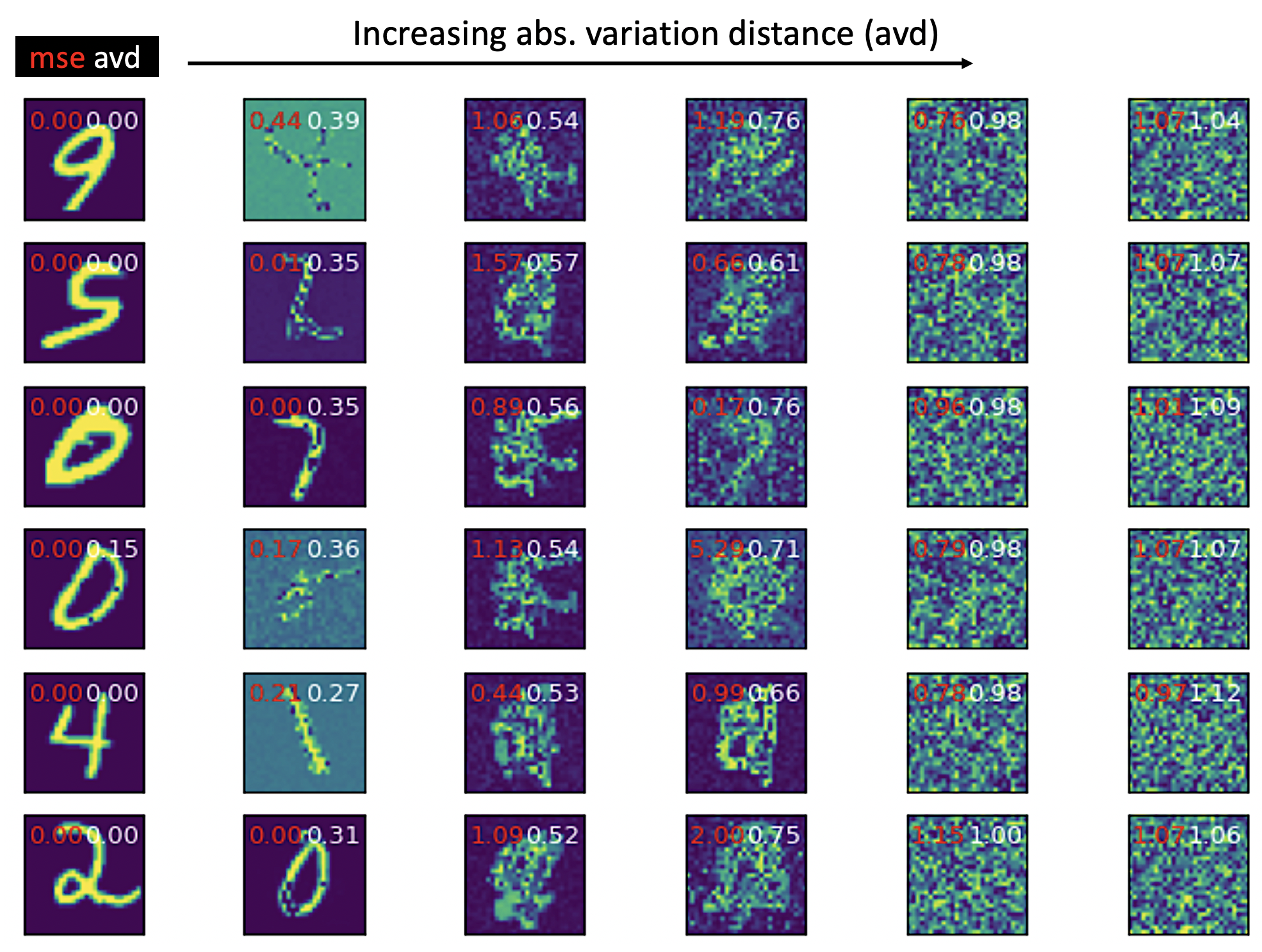}}
\subfigure[]{\includegraphics[trim={0.4cm 0.5cm 0cm 0.5cm},width=0.5\textwidth]{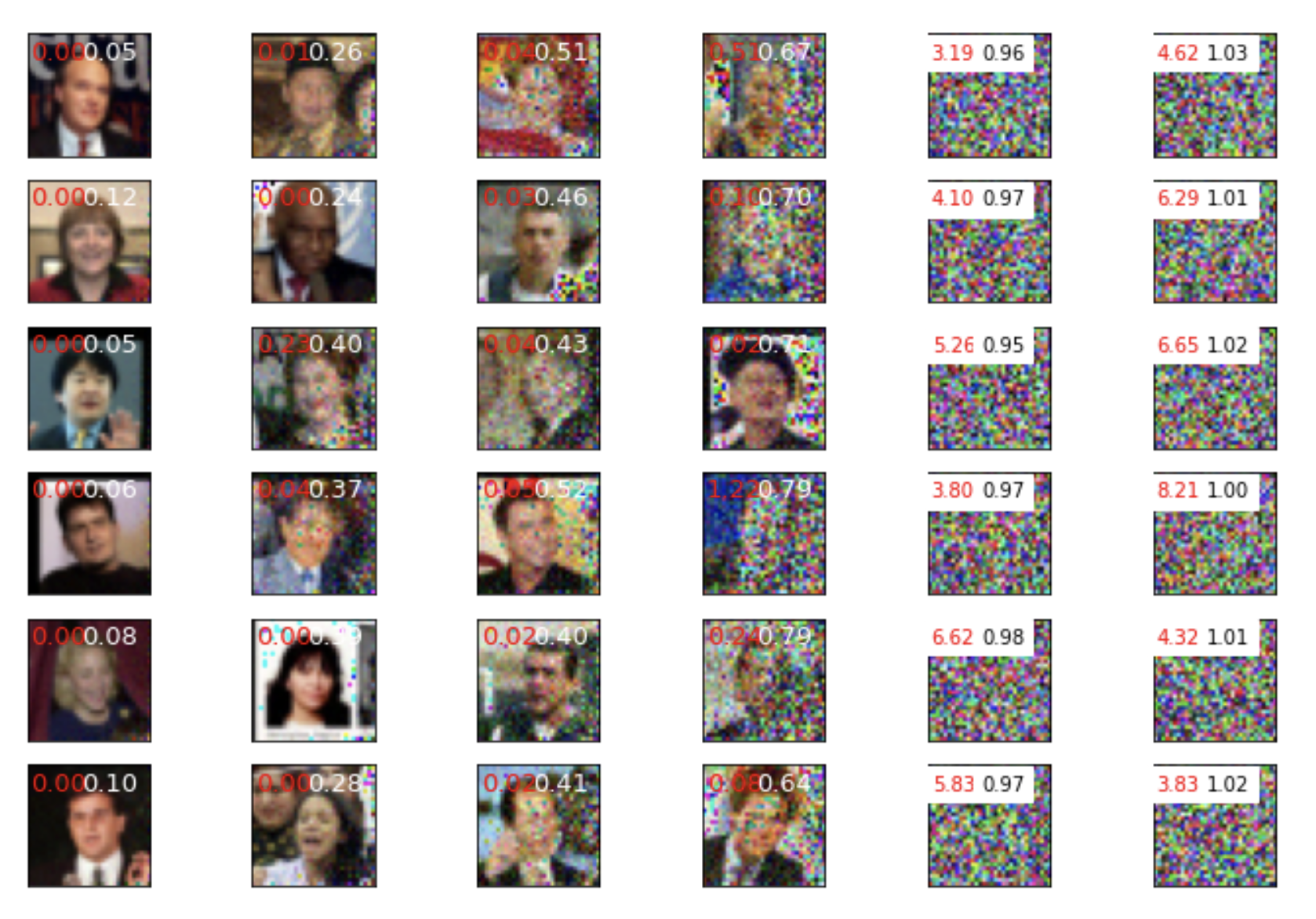}}
\centering
\caption{Random recovered vectors from MNIST (a) and LFW (b) datasets, column-wise sorted via the abs. variation distance measure. The AVD values were scaled to 1.0 by dividing the AVD for the distance between the initial uniform noise input image, to a black image.}
\label{avd}
\vskip -0.2in
\end{figure}

\subsection{Criterion for Successful Attack}

Many studies for improving attacks focus on fidelity of recovery and rate of convergence, e.g. \cite{zhao2020idlg,qian2021minimal,Yin21}. Our focus is the opposite, to determine if any information regarding the data can be recovered. Therefore our criterion for a successful attack is minimal and we define it as the ability to obtain information for one input vector in a batch. For example in MNIST this is the ability to recognise any information that is distinguished from noise. The mean square error (MSE) and structural similarity index measure (SSIM)~\cite{Wang04} are useful for high fidelity and small changes between images. We found that these metrics are not reliable and cannot be used as indicator for information leakage in datasets such as MNIST where the information is sparse. We can show this visible information in a random sample of recovered vectors from attacks in figure~\ref{avd}(a). The MSE indicator is not sufficient in the intermediate range where information is visible but noisy, blended, or there are other patterns that can significantly skew the results. 

A more suitable indicator is to compare the spatial gradient of the recovered image and source image:
\begin{equation}
    \mathrm{AVD}(v^{source},v^{target}) =||(|\nabla v^{source}_{x,y}|-|\nabla v^{target}_{x,y}|)||
\end{equation}
where $\nabla v=\frac{dv}{dx}+\frac{dv}{dy}$ is the pixel-wise gradient. 
The absolute variation distance allows to consider boundaries and edges in images which are a common discriminator in visual recognition, whilst the gradient of noise remains as noise. A random sample of attack results after 550 iterations of optimisation are shown in figure~\ref{avd}. A qualitative assessment of the results shows that recovery of data is more visible as AVD decreases in a continuous manner. In contrast the MSE metric for MNIST fluctuates drastically when the image is not completely clear or a blend, and can obtain various values similar or higher than the MSE for the pure noise input. Using this qualitative observation, we can define a threshold range between 0.6, where numbers are starting to emerge, and beyond 0.8, where information, at least for human eyes, is not visible. 

For the LFW dataset (figure~\ref{avd}), we see the same trend of good correlation between AVD and the revealed information, but the MSE metric is also a reasonable indicator as the images contain highly dense information. We use similar range of threshold of 0.6, and 0.8 to indicate a successful recovery.
\footnote{We note that using the variation by measuring the entropy, also yields in a very compatible metric, but was not used in this study. In that case, the relative information can be measured as:
\begin{equation}
\Delta S_{av}= -p_0\log |\nabla v^{source}_{x,y}|/|\nabla v^{target}_{x,y}|
\end{equation}
Here $p_0$ is used here to be the expected value of the initial input vector to the attack, which in our case is uniform noise (0,1), so $p_0=0.5$.}

\subsection{Datasets and Attack Experiments}

We conduct gradient inversion attack experiments on two representative datasets, MNIST Handwritten Digit \cite{lecun2010mnist} and Labelled Faces in the Wild (LFW) \cite{LFWTech}, to illustrate how our proposed defense strategies successfully minimise information leakage without performance degradation. These two dataset are commonly used among researchers to study attacks~\cite{Zhu19, zhao2020idlg, MelisSCS19, ShokriSSS17}. For each experiment we carried out 10-20 trials for each of the 3 attacks presented in table~\ref{attacks}. We analyse the recovery rate, which is the percentage of trials that lead to successful recovery. A successful recovery is determined by a threshold for the AVD metric (in MNIST) or MSE metric (in LFW) at the end of the every trial. In terms of the optimisation scheme, we utilized the standard optimisation scheme, LFBGS, with learning rate ($lr$) of $0.05$ and $550$ iterations for running a proxy model to attack.  We also carried out complementary tests with $1200$ iterations, and $lr$ of 0.025 to further showcase the validity of our results. The detailed analysis is available in the supporting information. 

To evaluate the performance of the neural networks we trained the LeNET and dense layer models using an SGD optimiser on the MNIST dataset for 60 epochs. We will release the source code to reproduce these results upon acceptance of the paper.

 
\section{Experimental Results}\label{sec:results}
\subsection{Single Dense Layer}\label{sec:linear}
A single layer attack is a valuable experiment to clearly demonstrate our strategy. As we have shown earlier we can recover the data directly from a batch without numerical optimisation. Here we explore the results from the attacks on a simplified single layer. In this case we look only at the MNIST dataset, as linear regression can provide a very useful though not optimal model for prediction. We explore different batch sizes, $B{=}1,2,4,6$ using the AVD metric with two thresholds to determine a successful recovery.
We calculate recovery rates for each experiment strategy, MSE vs. CEL as loss function, both followed by softmax and random drawing of labels vs. equal labels in a batch. 

\begin{figure}[ht]
\begin{center}
\leftline{\includegraphics[width=0.44\textwidth]{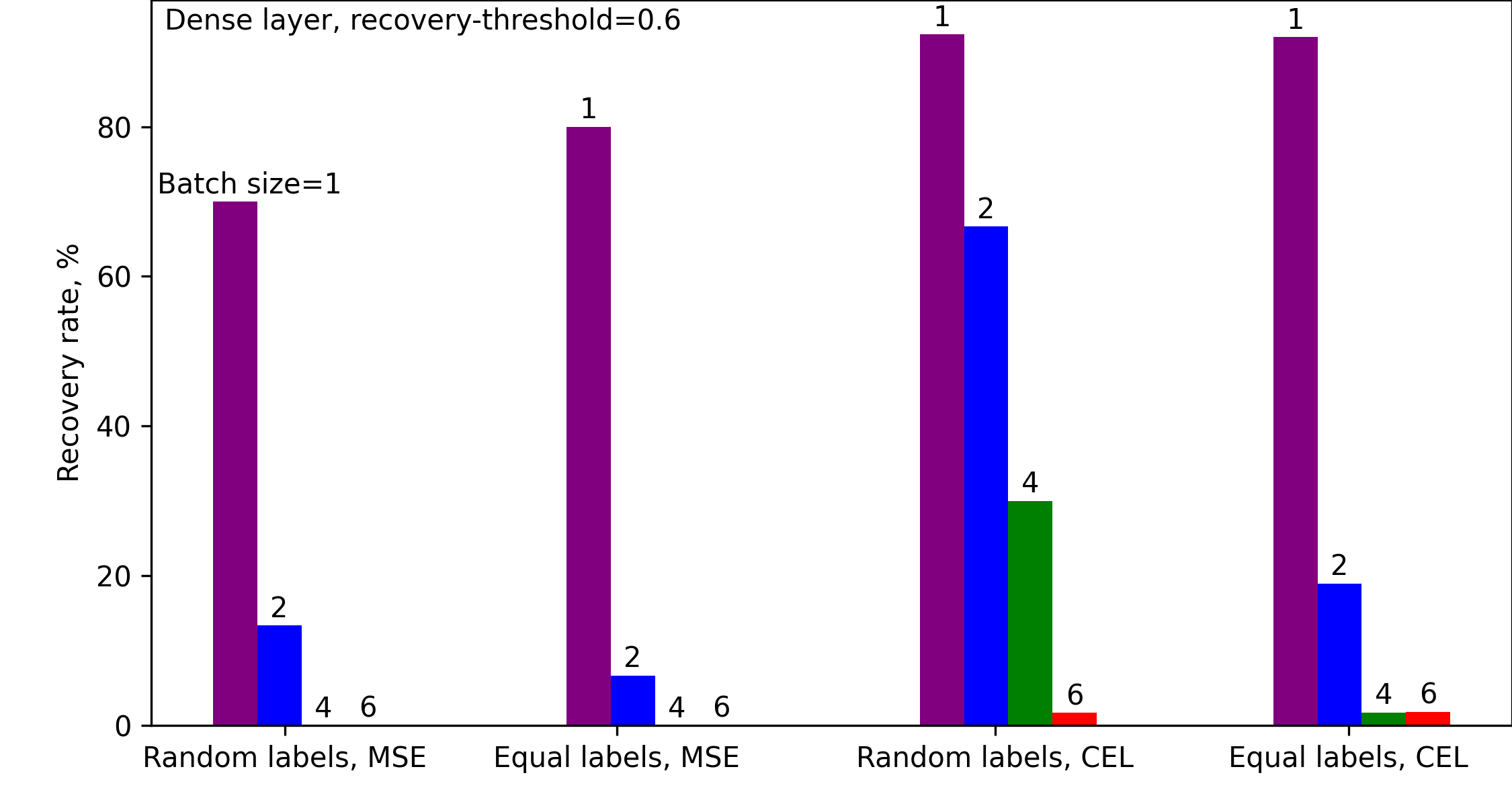}}
\leftline{\includegraphics[width=0.44\textwidth]{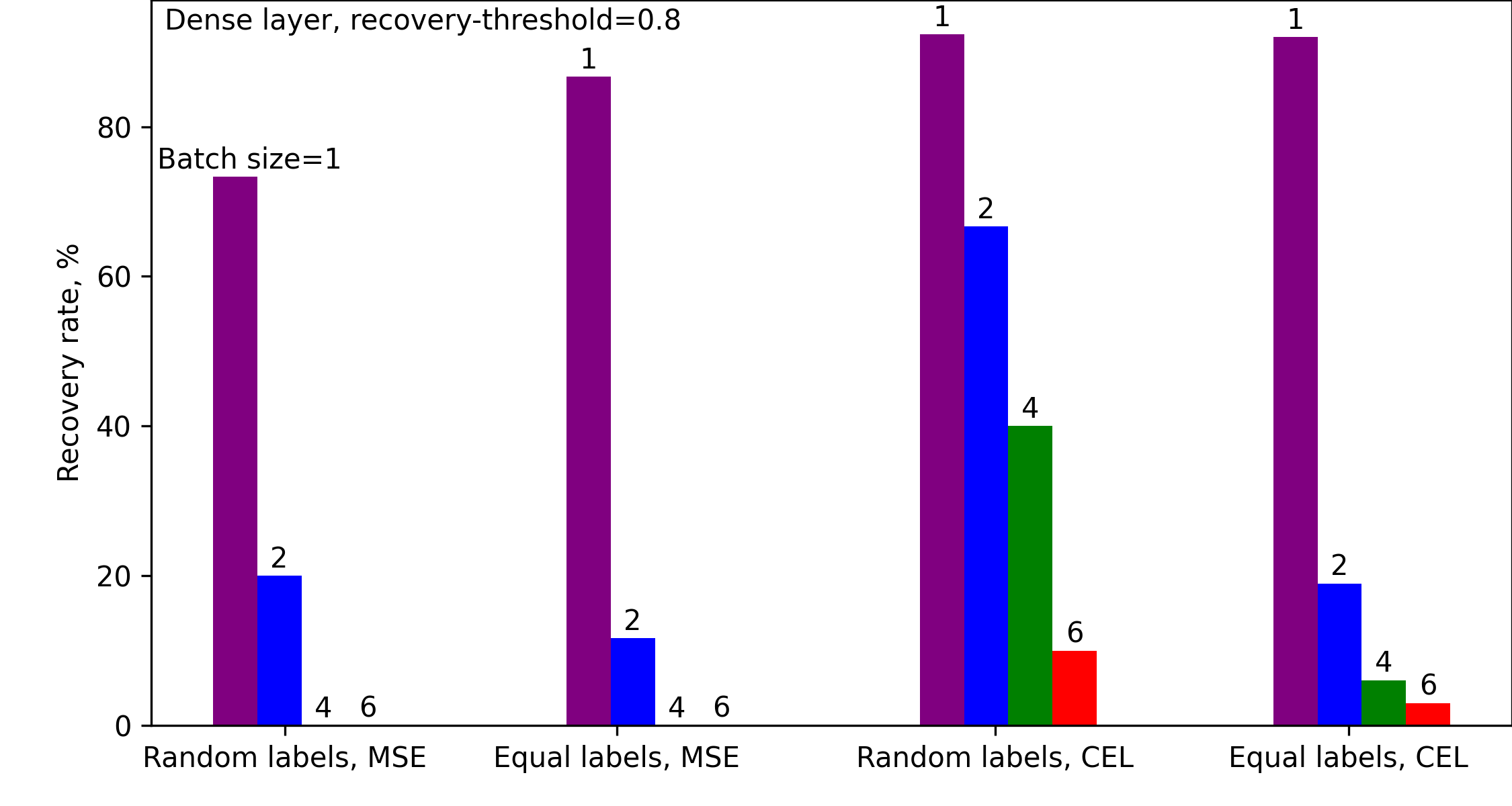}}

\caption{Recovery rates (in percents) for MNIST dataset using one linear dense layer. We preset the rate of recovery using the AVD metric with threshold 0.6 (a), and 0.8 (b). The bars show different mini-batch sizes, and the $x$-axis differentiates between MSE loss function, CEL function and the cases of drawing random and equal labels in CEL function.}
\label{mnist-linear}
\end{center}
\vskip -0.1in
\end{figure}

\begin{figure}[ht!]
\leftline{\includegraphics[clip,trim={0.0cm 0.3cm 0cm 0.3cm}, width=0.43\textwidth]{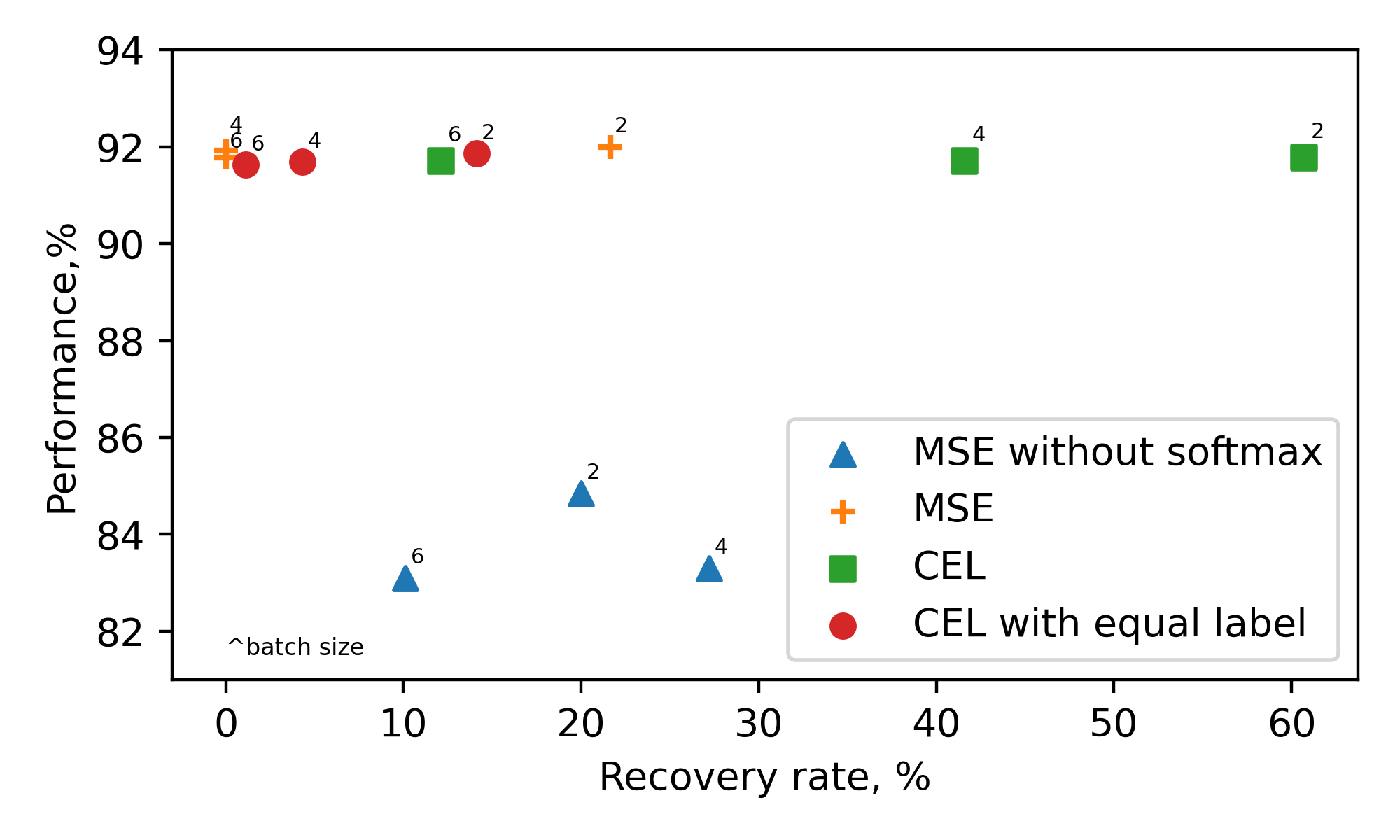}}
\caption{Performance of MNIST in single net architecture sizes at various strategies and at batch sizes 2, 4 and 6. The y-axis represents classification accuracy and the x-axis shows fraction of recovered images. The figure shows that recovery rate is high for cross entropy, and either mixing the labels or using a different loss function can reduce recovery rate without sacrificing the classification accuracy.}
\label{ratelinear}
\vskip -0.1in
\end{figure}

Figure~\ref{mnist-linear} shows that MSE and a batch of equal labels provides very low recovery rates for batches of size 4 and 6. This result is observed for both threshold values 0.6 and 0.8. The additional value of using MSE with equal labeling is minor. The regular approach of using random data in the batch with CEL is observed to be the most vulnerable. The importance of these results can be justified by looking at the training performance of these networks as presented in figure~\ref{ratelinear}.
It is shown that performance has in fact remained intact, even for MSE as an objective function followed by softmax, which is not typically applied in classification tasks, and also for equal labels despite the possible diversification issue that this may raise~\footnote{Hyper-parameters were not optimised and similar to all configurations.}. 

We note that in the aggregation of a central model, we update the model only after aggregation of the gradients from clients, so the diversification of labels in the batch of equal labels happens naturally. It therefore enables similar performance to the random label. We also observe that the MSE loss without softmax results in lower performance for a single dense layer. However we obtained the opposite behaviour for LeNET as we show next, so this discrepancy may be addressed by further optimisation through tuning the network hyper-parameters.

\subsection{Convolutional neural network recovery rates}
\begin{figure}[ht!]
\includegraphics[trim={0.3cm 3cm 0cm 0cm},width=0.5\textwidth]{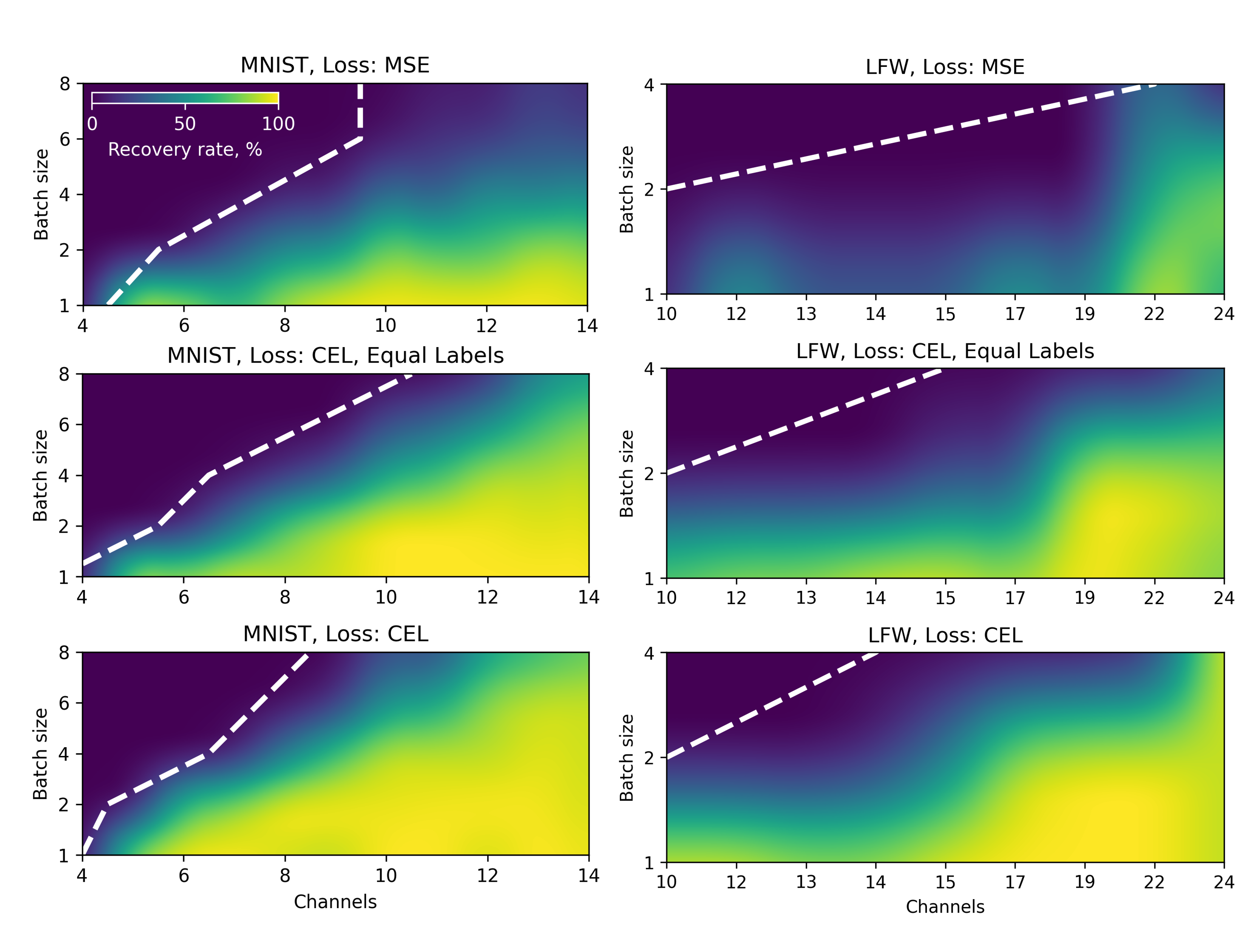}
\caption{Analysis of recovery attack success rate in MNIST and LFW with CNN model, LeNET architecture. The white dashed line represents the boundary of zero rate success recovery.}
\label{2d-cnn}
\vskip -0.1in
\end{figure}

Results for the linear layer in section~\ref{sec:linear} show the clear numerical challenge in inverting a batch while employing the suggested gradient mixing strategies. Here we explore how this affects the attack success in a widely used convolutional neural network for image recognition, LeNET~\cite{Lecun90}. We show a 2D map of the recovery rates for attack experiments on different batch sizes and channels. The results are presented in figure~\ref{2d-cnn} for the MNIST and the LFW dataset. The maps also show the boundaries for zero recovery rates. The trends show that overall the MSE and equal label strategies provide a useful decrease in recovery rates that permit a practitioner to deploy a larger model with data protection supported by the mixing of gradients. This effect is especially dominant for the LFW dataset. However we expect that with a much larger number of channels the data can be recovered solely by the convolutional layers. A detailed interplay between the minimal required channels in convolutional layers with or without dense layers was discussed recently by \cite{qian2021minimal}. 

\begin{figure}[ht!]
\centering
\leftline{\includegraphics[clip,trim={0.0cm 0.2cm 0cm 0.4cm}, width=0.47\textwidth]{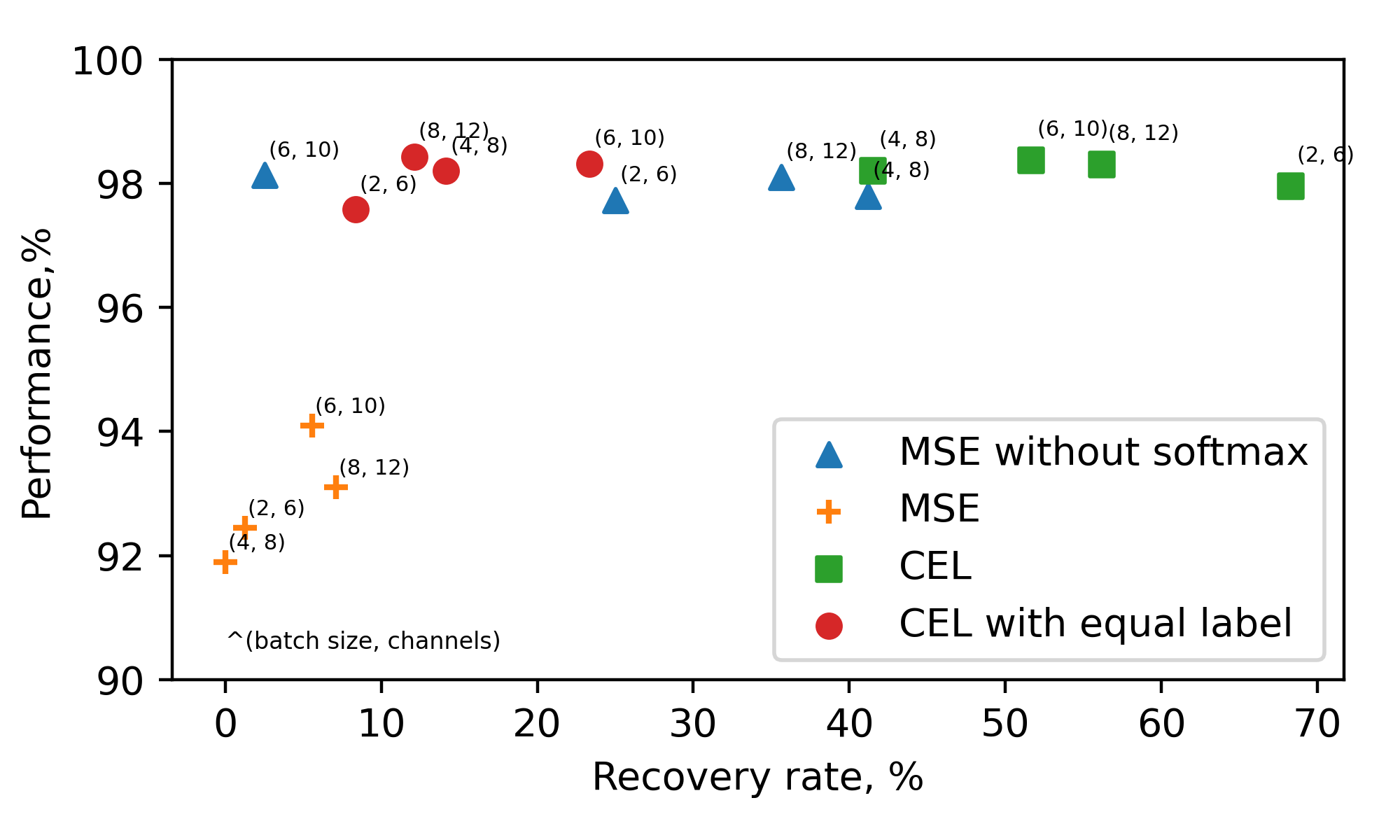}}
\caption{Performance of MNIST dataset on LeNET network architecture with different channels and batch sizes and for various gradient mixing strategies.}
\label{lenet-rate}
\vskip -0.1in
\end{figure}

Finally we carry out a performance test for LeNET for each strategy and with varying batch sizes and number of channels for the MNIST dataset. We show the comparison of the network performance against the recovery rates in figure~\ref{lenet-rate}. Results show that performance is also kept relatively intact allowing a clear benefit of privacy protection in comparison to the typical CEL and random label selection. Interestingly, in contrast to the single dense layer case, here MSE with softmax performs less well than other networks. Our results for LeNET show the benefit of choosing strategies for mixing gradients as, in many cases, the maximum batch size that can be used is limited (e.g. in distributed training over clients with sparse data).

\section{Conclusions}
We have shown that by simple architecture choices it is possible to prevent the recovery of data from widely used gradient inversion attacks. The choice of loss function and the drawing of equal labels in a batch results in strategic mixing of the gradients in typical neural networks with dense linear and convolutional layers. We revisited the analysis of the dense linear layer and have shown that mixing gradients makes a linear layer much less vulnerable to attack, and in fact, without mixing gradients, it is possible to recover directly all batch vectors due to the de-mixing nature of the typical cross entropy loss function. Our suggested strategies for mixing gradients maintain network performance which is in contrast to common methods that apply noise to the gradients and drop performance significantly. Additionally, in practice, one could combine the mixed gradients strategies further with noise or other defense methods to obtain an even more secure federated learning policy. Finally, we have introduced a new metric, absolute variation distance, to measure the relative information recovered by gradient inversion attacks. The AVD metric, which is derived from total variation, can distinguish information from noise especially for datasets that have sparse information such as the digit images in the MNIST dataset. 

We hope that this work prompts the development of new gradient mixing strategies that further secure data in distributed learning. Further work will also study the effect of more complex architectures and the creation of a privacy policy for federated learning that will allow a trustful federated learning platform.
\label{submission}

\bibliography{main}
\bibliographystyle{icml2021}

\end{document}